\documentclass{article}

\usepackage{microtype}
\usepackage{graphicx}
\usepackage{subcaption}
\usepackage{booktabs} % for professional tables

\usepackage{hyperref}

\usepackage[preprint]{icml2026}

\usepackage{amsmath}
\usepackage{amssymb}
\usepackage{mathtools}
\usepackage{amsthm}

\usepackage{wrapfig}
\usepackage{caption}
\usepackage{amsmath}
\usepackage{subcaption}
\usepackage{bbding}
\usepackage{graphicx}
\usepackage{multirow}

\usepackage[utf8]{inputenc}
\usepackage{subcaption}     % 子图支持
\usepackage{enumitem}       % 列表自定义
\usepackage{xcolor}         % 颜色支持
\usepackage[most]{tcolorbox} % 带皮肤库的彩色框

\newcommand{\ie}{\textit{i}.\textit{e}.}
\newcommand{\eg}{\textit{e}.\textit{g}.}

\AtEndPreamble{
    \usepackage[capitalize]{cleveref}
    \crefname{section}{Sec.}{Secs.}
    \Crefname{section}{Section}{Sections}
    \Crefname{table}{Table}{Tables}
    \crefname{table}{Tab.}{Tabs.}
}
\usepackage{colortbl}
\definecolor{mygray}{gray}{0.95}
\definecolor{my_green}{RGB}{82,208,80}
\usepackage{enumerate}
\usepackage{makecell}
\definecolor{00red}{RGB}{236,35,35}

\usepackage[capitalize,noabbrev]{cleveref}

\theoremstyle{plain}

\theoremstyle{definition}

\theoremstyle{remark}

\usepackage[textsize=tiny]{todonotes}

\icmltitlerunning{MM-DeepResearch: A Simple and Effective Multimodal Agentic Search Baseline}

\newcommand\blfootnote[1]{%
  \begingroup
  \renewcommand\thefootnote{}\footnote{#1}%
  \addtocounter{footnote}{-1}%
  \endgroup
}

\begin{document}

\twocolumn[
  \icmltitle{MM-DeepResearch: A Simple and Effective Multimodal Agentic Search Baseline}



  \icmlsetsymbol{equal}{*}

  \begin{icmlauthorlist}
    \icmlauthor{Huanjin Yao$^{1}$}{}
    \icmlauthor{Qixiang Yin$^{3}$}{}
    \icmlauthor{Min Yang$^{1, \text{\Envelope}}$}{}
    \icmlauthor{Ziwang Zhao$^{1}$}{}
    \icmlauthor{Yibo Wang$^{4}$}{}
    \icmlauthor{Haotian Luo$^{5}$}{} \\
    \icmlauthor{Jingyi Zhang$^{4}$}{}
    \icmlauthor{Jiaxing Huang$^{2, \text{\Envelope}}$}{} \\

  \end{icmlauthorlist}



  \icmlkeywords{Machine Learning, ICML}

  \vskip 0.3in
]

\blfootnote{
$\text{\Envelope}$ Corresponding author. 
$^1$ ByteDance;
$^2$ Hong Kong Polytechnic University;
$^3$ Zhongguancun Academy;
$^4$ Nanyang Technological University;
$^5$ Sichuan University.
} 

% Huanjin Yao, Qixiang Yin, Min Yang, Ziwang Zhao, Yibo Wang, Jingyi Zhang, Jiaxing Huang, Jiaya Jia




\begin{abstract}
We aim to develop a multimodal research agent capable of explicit reasoning and planning, multi-tool invocation, and cross-modal information synthesis, enabling it to conduct deep research tasks.
However, we observe three main challenges in developing such agents: (1) scarcity of search-intensive multimodal QA data, (2) lack of effective search trajectories, and (3) prohibitive cost of training with online search APIs.
To tackle them, we first propose \textbf{Hyper-Search}, a hypergraph-based QA generation method that models and connects visual and textual nodes within and across modalities, enabling to generate search-intensive multimodal QA pairs that require invoking various search tools to solve.
Second, we introduce \textbf{DR-TTS}, which first decomposes search-involved tasks into several categories according to search tool types, and respectively optimize specialized search tool experts for each tool. It then recomposes tool experts to jointly explore search trajectories via tree search, producing trajectories that successfully solve complex tasks using various search tools.
Third, we build an offline search engine supporting multiple search tools, enabling agentic reinforcement learning without using costly online search APIs.
With the three designs, we develop \textbf{MM-DeepResearch}, a powerful multimodal deep research agent, and extensive results shows its superiority across benchmarks.
Code is available at \url{https://github.com/HJYao00/MM-DeepResearch}
\end{abstract}

\section{Introduction}
\label{Sec 1: Introduction}
Recently, reasoning Multimodal Large Language Models (MLLMs)~\cite{openai-o1, comanici2025gemini-2.5} have shown remarkable capabilities in solving complex tasks by explicitly generating intermediate reasoning processes.
Despite these advances, existing models remain fundamentally constrained by capacity-limited parameters with fixed and bounded knowledge, which restrict their ability to handle questions that go beyond their intrinsic knowledge, especially for information-intensive and open-world tasks.

To mitigate this limitation, early methods primarily adopted pre-defined information retrieval workflows, such as Retrieval-Augmented Generation (RAG)~\cite{chen2024mllm-rag} or prompt-based approaches~\cite{jiang2024mmsearch}, which retrieve external information as context in a pre-defined pipeline and then start reasoning upon the additional context.
However, these approaches decouple retrieval from reasoning processes, lacking the ability to iteratively adapt retrieval strategies as the model’s reasoning state evolves~\nocite{zhao2025pyvision,shang2025rstar2,zhang2025toolr1}, resulting in limited search capability and generalization.

\begin{figure}[t]
\centering
\includegraphics[width=1\linewidth]{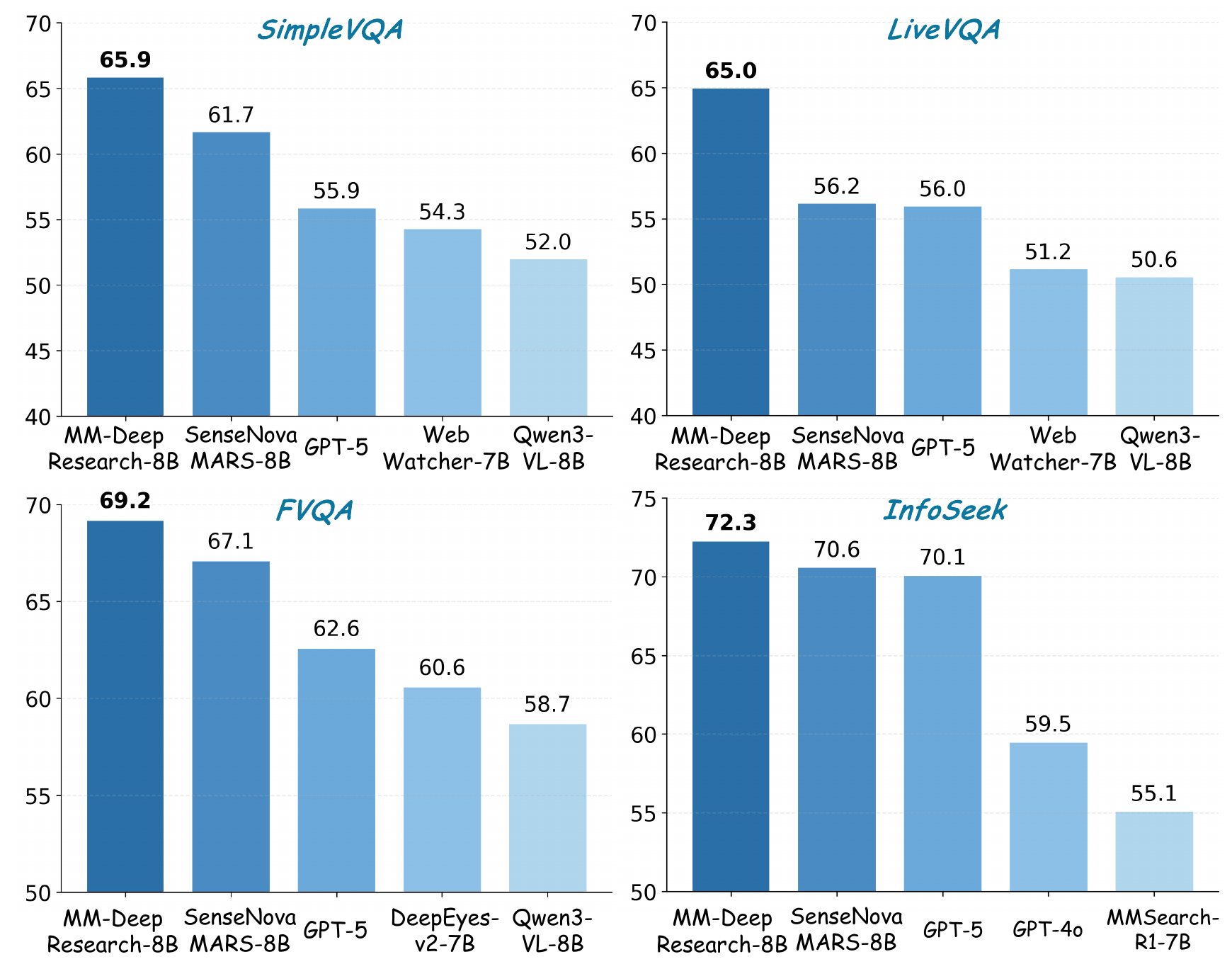}
\caption{
Overall performance of MM-DeepResearch-8B compares to other models across four benchmarks.
}
\label{fig:1}
% \vskip -0.25in
\end{figure}

We aim to develop a multimodal deep research model equipped with agentic search capabilities, including explicit reasoning and planning, multi-tool invocation, and cross-modal information synthesis, enabling it to conduct deep research tasks.
We observe several fundamental challenges in developing such agents:
(1) \textit{Scarcity of search-intensive multimodal QA data.} Publicly available search-intensive multimodal QA datasets with multi-turn search and multi-tool invocation are very limited, resulting in insufficient supervision to effectively incentivize agentic search capabilities in MLLMs.
(2) \textit{Lack of effective search trajectories.} Traditional prompt-based search trajectory synthesis approaches are primarily designed for single-turn search, making them inadequate for multi-turn scenarios that require multi-round interaction with various search tools during iterative reasoning.
(3) \textit{Prohibitive cost of training with online search APIs.} Most current methods rely on online search APIs (e.g., SerpAPI and Jina) during training, which can easily incur thousands of dollars per training run, substantially limiting extensive experiments and systematic exploration.

To address these challenges, we propose three complementary techniques: Hyper-Search, DR-TTS, and an offline multimodal search engine.
First, we propose \textbf{Hyper-Search}, a hypergraph-based method that models and connects visual and textual nodes within and across modalities, enabling to generate search-intensive multimodal QA pairs that require invoking various search tools to solve.
Specifically, Hyper-Search jointly models web visual and textual content as nodes, and captures the native relationships among these nodes via hyperedges, enabling a structured representation of complex multi-source and multimodal information dependencies.
Based on this hypergraph, Hyper-Search constructs search-intensive multimodal QA pairs by involving two or more hypergraph nodes across one or more hyperedges, ensuring that each generated question inherently requires multi-turn and multi-type search tool invocation to solve.

Second, we introduce Decompose–Recompose Tool Tree Search (\textbf{DR-TTS}) to effectively synthesis of search-intensive reasoning trajectories that utilize various search tools to complete tasks.
DR-TTS first decomposes search-involved tasks into several categories according to search tool types, and respectively train specialized search tool experts via reinforcement learning, with each expert mastering a single tool.
% This design simplifies training process and enhances per-tool proficiency.
This decomposition design simplifies learning complexity of search tools and enhances per-tool proficiency.
DR-TTS then recomposes the tool experts to jointly explore and identify valid search trajectories via tree search, producing high-quality search trajectories for SFT.
Compared to directly generating trajectories with a single model, DR-TTS mitigates tool-use bias and enhances per-tool proficiency, thus achieving more balanced and better exploration within and across tools, and significantly increasing trajectory exploration diversity and success rates.

Third, upon above two designs, we construct an offline search engine that supports both information-based and knowledge-based search to handle a wide range of complex real-world tasks.
The information-based search tools retrieve visual and textual context from pre-constructed multimodal corpus to support search-intensive tasks, while the knowledge-based search tool accepts complex search queries and return knowledge-intensive content generated by models.
Compared with online search engines, our offline search engine provides faster tool responses and reduces tool-interactive training time.
More importantly, it circumvents the expensive online search APIs cost, which can amount to thousands of dollars per training run.

Building on these designs, we train a powerful multimodal deep research agent named \textbf{MM-DeepResearch}.
Trained with the search-intensive QA data generated by Hyper-Search, the search tool trajectories synthesized via DR-TTS, and the offline search engine, MM-DeepResearch achieves superior performance across multiple benchmarks when evaluated using online search APIs, outperforming previous agents trained with costly online search APIs.

The main contributions of this work are fourfold.
First, we propose Hyper-Search that introduces hypergraph to model and connect nodes within and across modalities, enabling to generate high-quality search-intensive multimodal QA pairs.
To the best of our knowledge, this is the first work that introduces the concept of hypergraph for search-intensive QA data synthesis.
Second, we design DR-TTS, which decomposes search-involved tasks to enable training specialized search tool experts, and recomposes tool experts to jointly explore and synthesize high-quality search trajectories.
Third, we build an offline search engine with multiple search tools, enabling agentic reinforcement learning without using costly online search APIs.
Fourth, extensive experiments demonstrate the superiority of our approaches and models across various multimodal deep research tasks.

\section{Related Work}
\label{Sec 2: Related work}

\subsection{Agentic MLLMs}
Early efforts leverage large-scale multimodal pretraining and instruction tuning to build general-purpose MLLMs for vision-language understanding tasks~\cite{llava1.5, an2025llavaonevision15fullyopenframework}, and adopt post-training techniques to incentivize long-chain reasoning capabilities for complex reasoning tasks~\cite{huang2025visionr1, zhang2025r1vl}.
Recent work further emphasizes agentic behaviors~\cite{yao2025agentic_mllm_survey}, enabling MLLMs to autonomously and iteratively select and invoke tools for more complex tasks.
Recent agentic foundational MLLMs~\cite{vteam2026glm45vglm41vthinkingversatilemultimodal}, such as Qwen3-VL~\cite{Qwen3-VL}, are equipped with native tool-use capabilities by including diverse tool-interaction trajectories in agentic pretraining.
In this paper, we aim to develop a multimodal deep research agent with improved capabilities in search tool invocation and multimodal information synthesis.

\subsection{Workflow-based Search Agents}
Workflow-based search agents employ predefined and static information-seeking pipelines to retrieve external knowledge for reasoning. These approaches can be broadly categorized into RAG-based and prompt-based methods.
(1) RAG-based approaches~\cite{wang2025vragrl,li2025towards_rag} augment MLLMs by retrieving external knowledge from databases using similarity-based retrieval and injecting the retrieved content into the model input.
(2) Prompt-based methods~\cite{zheng2025deepresearcher,wu2025agentic_reasoning} explicitly orchestrate search behaviors through hand-crafted workflows encoded in prompts, where tool usage patterns are predefined during inference.
These methods improve factual grounding and performance on knowledge-intensive tasks by incorporating external knowledge.
However, workflow-based search agents decouple search from reasoning and are constrained by fixed and rigid pipelines, making them prone to insufficient or excessive search and consequently limiting their generalization performance.
These limitations motivate us to explore multimodal deep research agents that actively and iteratively invoke search tools in a tightly coupled reasoning–search manner.

\subsection{Multimodal Deep Research Agents}
Recent work has introduced deep research agents that equip LLMs with agentic search capabilities for iterative information seeking and evidence synthesis through interaction with external text search tools, represented by OpenAI DeepResearch~\cite{openai_deep_research} and MiroThinker~\cite{miromind2025mirothinker}.
Beyond text-only agents, several prior works~\cite{yang2025multimodal_deepresearcher,liu2025visual_Agentic,xiao2025m2io_r1} extend deep research paradigms to multimodal domains.
MMSearch-R1~\cite{wu2025mmsearchr1} first employs end-to-end RL to equip MLLMs with the ability to use image and text search tools.
WebWatcher~\cite{geng2025webwatcher} synthesizes search-intensive QA pairs and converts them into VQA data for training multimodal deep research agents.
However, most existing works remain closed-source with respect to search-intensive VQA datasets and search tool trajectories, and they rely heavily on online search APIs, posing significant challenges for the development of multimodal deep research agents.
% close model / data, poor performance, 
To tackle these challenges, this paper proposes Hyper-Search for generating multimodal search-intensive QA data, DR-TTS for synthesizing search trajectories, and an offline search engine for RL. Together, these components enable us to train a deep research agent and achieve strong empirical performance.

\section{MM-DeepResearch Data}
In this section, we focus on the data-level challenges of multimodal deep research.
We first present a hypergraph-based method (Hyper-Search) for search intensive QA construction in \cref{Sec 3.1: hyper_search}.
Then, we propose DR-TTS to jointly explore and find valid search tool trajectories in \cref{Sec 3.2: DR-TTS}.

\subsection{Hyper-Search for QA Data Generation} 
\label{Sec 3.1: hyper_search}
\begin{figure*}[t]
\centering
\includegraphics[width=0.9\linewidth]{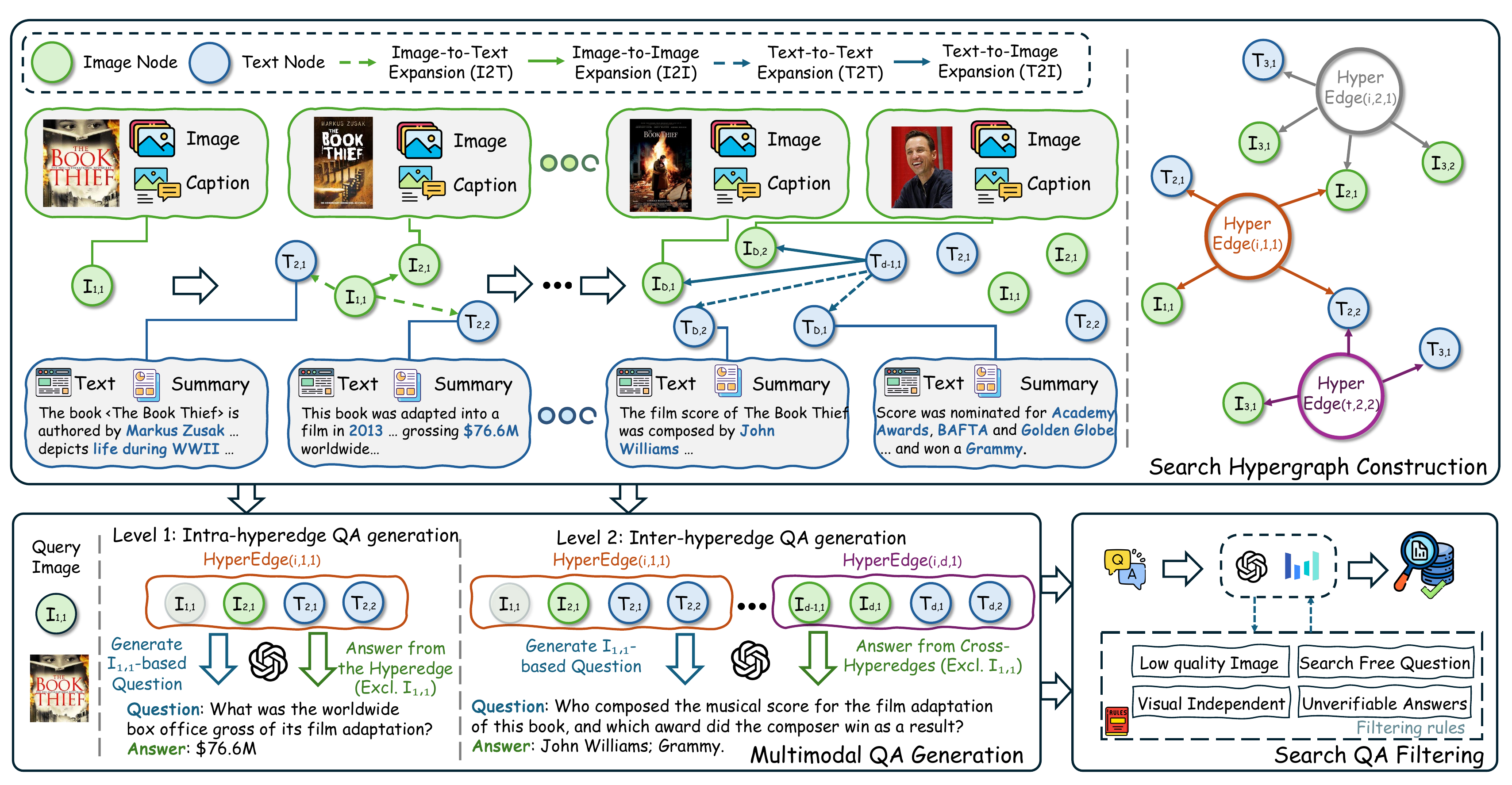}
\caption{
Overview of Hyper-Search for generating search-intensive QA data via hypergraph construction, QA generation, and filtering.
}
\label{fig:2}
\vskip -0.1in
\end{figure*}

As shown in~\cref{fig:2}, the construction of multimodal search-intensive QA data via Hyper-Search consists of three stages:
(1) Hypergraph construction, which models web images, webpage content, and their relationships;
(2) Multimodal QA generation, where QA pairs are generated based on the constructed hypergraph; and
(3) Data filtering, which aims to remove low-quality and search-free multimodal QA data.

\subsubsection{Search Hypergraph Construction}
% % a hypergraph-based method for search-intensive QA construction that models and connects nodes within and across modalities, enabling to synthesize QA data for complex real-world tasks that require invoking various search tools to solve.

% Hyper-Search aims to build a hypergraph that models web information within and across modalities, capturing cross-source relationships grounded in the real-world web.
% This structured representation enables the synthesis of multimodal QA data for complex real-world tasks that require multiple interactions with various search tools to be effectively addressed.

\textbf{Node Definition.}
In Hyper-Search, we model web information using a hypergraph composed of two fundamental node types, \ie, image nodes $\mathcal{I}$ and text nodes $\mathcal{T}$.
Each image node $i_{d,k} \in \mathcal{I}$ corresponds to an image collected from an online source, while each text node $t_{d,k} \in \mathcal{T}$ represents the full content of a webpage, where $d$ denotes the expansion depth and $k$ indexes the nodes at that depth.
For each node, we leverage an MLLM to generate captions for image nodes and summaries for text nodes, which serve as informative cues during subsequent multimodal QA generation.
These representations enable modality-specific node expansion and interconnection, supporting the synthesis of multimodal QA pairs grounded in diverse information sources.

\textbf{Node Expansion.}
We design distinct expansion strategies for text and image nodes to retrieve additional relevant information and enrich the hypergraph.
(a) \textit{Image node expansion.} For each image node $i_{d,k}$, we define two expansion operations that respectively introduce new text and image nodes.
The first operation, image reverse search, queries the image to retrieve $K$ relevant webpage contents, yielding text nodes $\{t_{d+1,1}^{i_{d,k}}, \ldots, t_{d+1,K}^{i_{d,k}}\}$.
The second operation, image visual search, returns $K$ visually similar images, producing image nodes $\{i_{d+1,1}^{i_{d,k}}, \ldots, i_{d+1,K}^{i_{d,k}}\}$.
% Text Nodes Expansion
(b) \textit{Text node expansion.} For each text node $t_{d,k}$, we likewise design two expansion operations to extend the hypergraph with both text and image nodes.
For text node expansion, we use an MLLM to extract the top-$K$ informative webpage URLs from the native page content, each introducing a new text node corresponding to the retrieved webpage $\{t_{d+1,1}^{t_{d,k}}, \ldots, t_{d+1,K}^{t_{d,k}}\}$.
For image node expansion, in a similar manner, the LLM identifies the top-$K$ relevant image links directly from the native webpage content, which are used to form new image nodes $\{i_{d+1,1}^{t_{d,k}}, \ldots, i_{d+1,K}^{t_{d,k}}\}$.
The superscript denotes the parent node from which the new node is expanded.

\textbf{Hyperedge Connection.}
Each node expansion induces a hyperedge $e_{d,k}$ that connects the parent node with all newly generated nodes, capturing multimodal interactions.
For example, given an image node \(i_{d,k}\), its expansion produces new image nodes
$\{i_{d+1,1}^{i_{d,k}}, \ldots, i_{d+1,K}^{i_{d,k}}\}$ and text nodes
$\{t_{d+1,1}^{i_{d,k}}, \ldots, t_{d+1,K}^{i_{d,k}}\}$.
These nodes are grouped into a hyperedge defined as
$e_{d,k} = \{\, i_{d,k}^{i_{d-1,k}},\;
i_{d+1,1}^{i_{d,k}}, \ldots, i_{d+1,K}^{i_{d,k}},\;
t_{d+1,1}^{i_{d,k}}, \ldots, t_{d+1,K}^{i_{d,k}} \,\}.$
This construction explicitly models the cross-modal retrieval relationships among the connected nodes.
% Hyperedges for text nodes are constructed analogously.

\textbf{HyperGraph.} We initialize hypergraph construction from an image node. Through continual expansion, Hyper-Search incrementally constructs a depth-$D$ hypergraph that organizes multimodal web information and provides structured relational units for downstream multimodal QA generation.

\subsubsection{Multimodal Search QA Generation}
QA generation is conducted at the hyperedge level, leveraging cross-modal and cross-source information within and across hyperedges to generate multimodal, search-intensive QA data.
Based on the scope of aggregated hyperedge, we define two generation strategies: \emph{intra-hyperedge} and \emph{inter-hyperedge} QA generation.

% \textbf{Level 1: Intra-hyperedge QA generation.}
% At Level~1, QA generation is performed within a single hyperedge \(e_{d,k}\).
% Specifically, we select an image node \(i_{d,k}\) as the query image and generate a QA data conditioned on the remaining image captions and webpage contents contained in \(e_{d,k}\).
% The generated question is explicitly required to be vision-dependent based on \(i_{d,k}\) and search-intensive, while the corresponding answer must be come from the provided evidence.
% Since answers span multiple modalities, data at this level can effectively incentivize the model to actively search using both image and text tools.

\textbf{Level 1: Intra-hyperedge QA Generation.}
At Level~1, QA generation is performed within a single hyperedge \(e_{d,k}\).
Specifically, we select an image node \(i_{d,k}\) as the query image and generate QA data conditioned on the remaining image captions and webpage content as evidence within \(e_{d,k}\).
The generated search-intensive questions are explicitly vision-dependent on \(i_{d,k}\), while the corresponding answers must be derived from the provided evidence.
As the evidence span multiple modalities, this level of data effectively encourages the model to actively search using both image and text tools.

\textbf{Level 2: Inter-hyperedge QA generation.}
At Level~2, we broaden the evidence scope to increase search depth and difficulty, where QA generation integrates information across multiple hyperedges \(\{e_{d_1,k_1}, \ldots, e_{d_m,k_m}\}\).
Specifically, an image node is selected as the query image, and the QA data is generated based on aggregated images and webpage content drawn from these hyperedges.
This level typically requires multiple rounds of search, thereby encouraging agentic search behaviors that demand deeper exploration and more extensive information aggregation.

\subsubsection{Search QA Filtering}
To ensure data quality, we use MLLMs (\ie, GPT-5~\cite{GPT-5} and Seed1.8~\cite{seed1.8}) to filter low-quality multimodal QA instances.
This process removes low-quality images, duplicated QA pairs, visually irrelevant or search-free questions, and answers unverifiable from the evidence.
The resulting dataset contains 3K search-intensive multimodal QA pairs, termed \textbf{Hyper-Search-3K}.

\begin{figure*}[t]
\centering
\includegraphics[width=1\linewidth]{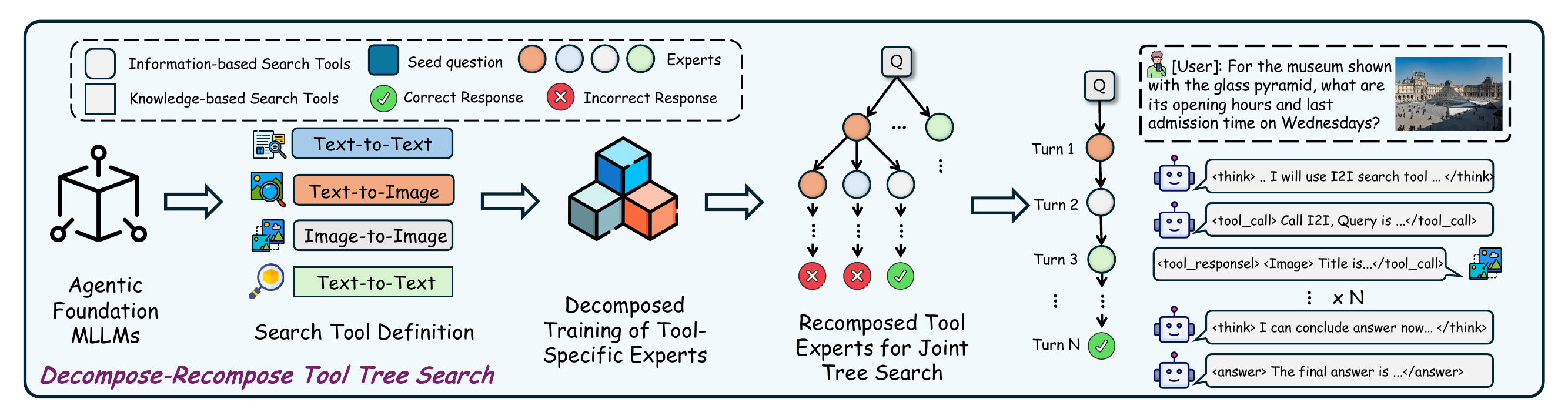}
\caption{
Overview of Decompose–Recompose Tool Tree Search for exploring search trajectories.}
\label{fig:3}
\vskip -0.1in
\end{figure*}

% reasoning trajectories (with/using tools)
\subsection{DR-TTS for Search Trajectory Synthesis}
\label{Sec 3.2: DR-TTS}
We propose Decompose–Recompose Tool Tree Search (DR-TTS), a tree search method for synthesizing search tool trajectories.
DR-TTS first categorizes search-related tasks according to required search tools and decomposes model training to optimize specialized experts for each tool respectively.
It then recomposes these tool experts to perform joint tree search, aiming to discover valid trajectories with higher exploration success rates and greater trajectory diversity.

\textbf{Search Tool Definition.}
We define information-based and knowledge-based search tools, which enable agents to acquire grounded factual information and specialized knowledge for more accurate trajectory search.
\textit{(1) Information-based search tools} retrieve real-world information by matching multimodal queries against external web sources.
We design three types of information-based search tools (\ie, text-to-text, text-to-image, and image-to-image retrieval) to obtain required textual and visual information based on different query types.
Specifically, text-to-text retrieval returns relevant webpages given a textual query; text-to-image retrieval returns relevant web images based on a textual query; and image-to-image retrieval returns visually similar images given an image query.
These tools provide grounded textual and visual inforamtion, enabling the model to address search-intensive reasoning tasks.
\textit{(2) Knowledge-based search tools} are designed to provide knowledge-intensive information by querying language models with complex queries.
We formulate these tools as text-to-text interactions, in which our agents pose natural language queries and receive responses generated by domain-specific expert models.
Such tools are useful for domain-specific reasoning tasks that require information not directly accessible through web retrieval.

\textbf{Decomposed Training of Tool-Specific Experts.}
Directly performing tree search with a pre-trained agentic foundational MLLM often leads to suboptimal performance, as the model frequently produces invalid tool calls and exhibits limited capability in integrating retrieved information, significantly hindering effective exploration.
To address this issue, we first categorize search-related tasks according to the required search tools and decompose model training to optimize specialized experts for each tool.
Specifically, we classify search-related training data by tool type using GPT.
Each tool-specific subset is then used to train a corresponding expert via our RL strategy described in~\cref{sec 4.2: rl}, without cold start.
This process results in a set of \(M\) specialized tool experts, each proficient in a single search tool, reducing learning complexity and enhancing per-tool proficiency.

\textbf{Recomposed Tool Experts for Joint Tree Search.}
We then recompose these tool experts to collaboratively explore and find search trajectories that successfully solve the tasks via tree search.
Starting from a root node (\ie, input question $Q=\{T, I\}$), we expand $M$ child nodes, each generated by one of the tool experts.
Each node in the tree, denoted as $s_i^j$, consists of a reasoning and planning step $t_i^j$, a tool call $c_i^j$, and the tool response $r_i^j$, where $i$ denotes the depth of the node and $j$ indexes the tool expert.
By branching over different tool experts at each depth, the search proceeds recursively across successive tree levels, exploring diverse expert-driven decisions and constructing candidate trajectories.
When an expert chooses to produce a final answer $a$ without invoking a tool, we evaluate the answer’s correctness using an LLM.
If the answer is verified to be correct, the entire tree search terminates; otherwise, nodes that produce incorrect final answers are treated as terminal and pruned from further expansion, while the search continues over the remaining nodes.

\textbf{Search Trajectory Extraction.}
For search trajectories that successfully produce a correct answer, we extract the corresponding explored path from the tree and construct trajectory data for supervised fine-tuning (SFT).
Each trajectory is represented as an ordered sequence of reasoning steps and tool interactions as $\tau = \{t_1, c_1, r_1, t_2, c_2, r_2, ..., t_n, a_n\}$. 
% where \(t_i\) denotes the reasoning and planning step, \(c_i\) the tool call, \(r_i\) the corresponding tool response at step \(i\), and \(a\) the final answer.
Through DR-TTS, tree-structured search over diverse tool experts enables the discovery of effective reasoning trajectories by balancing tool usage during exploration.
Ultimately, we collect 10K SFT trajectories using DR-TTS, referred to as DR-TTS-10K.

\section{MM-DeepResearch Agents}
Using the generated QA data and search tool trajectories, we train our MM-DeepResearch with a two-stage training recipe: (i) SFT on search tool trajectories as a cold start, and (ii) agentic multi-turn RL using offline search engine.

\subsection{SFT with Multi-turn Search Tool Invocation}
Using search trajectories explored by DR-TTS, we perform multi-turn SFT as a cold start, training the model to learn tool call patterns and boost cross-modal information integration capabilities.
The training objective maximizes likelihood of model-generated reasoning and planning steps $t_i$, search tool calls $c_i$, and the final answer $a_n$, while masking tool responses $r_i$ from the loss. The SFT objective is:
\begin{align}
\mathcal{L}_{\mathrm{SFT}} = - \mathbb{E}_{\tau \sim \mathcal{D}} 
\sum_{i=1}^{n} 
\log p_\theta\!\left(t_i, c_i, a_n \,\middle|\, Q, t_{<i}, c_{<i}, r_{<i}\right)
\end{align}
where the final answer $a_n$ is generated and supervised only at the final step, and tool responses $r_i$ are treated as observations conditioning the next-step prediction.
In this way, the model learns appropriate tool-invocation patterns and improves long-context information synthesis.

\subsection{Reinforcement Learning with offline Search Engine}
\label{sec 4.2: rl}
Unlike prior methods that use online search APIs during RL, which can be prohibitively costly, especially for GRPO~\cite{guo2025deepseek_r1} requiring multiple rollouts, we build an offline search engine to simulate real-world search environments and support both visual and textual retrieval.

\textbf{Offline Search Engine Construction.}
To enable the offline search engine to invoke the tools defined in~\cref{Sec 3.2: DR-TTS} during RL, we pre-collect a large-scale multimodal corpus comprising diverse images and texts.
Specifically, for textual information, we use GPT to generate multiple candidate search queries and pre-fetch the corresponding webpage content. To further expand the corpus, we additionally incorporate Wikipedia data.
For image-related information, we similarly pre-collect potentially relevant images and index them using a multimodal dense retrieval model.
This corpus construction substantially reduces search API costs by avoiding repeated online API calls during training, while still supporting effective learning of agentic search and tool-invocation behaviors.
Details of the offline corpus construction and retrieval implementation are provided in the appendix.

\textbf{Multi-Turn Reinforcement Learning.} To further incentivize model's deep research capabilities, we optimize it using multi-turn RL (\ie, GRPO) with various tools.
Specifically, given a question $Q$, the policy model generates a group of multi-turn search tool trajectories $\{\tau_1, \tau_2, \ldots, \tau_G\}$.
% These trajectories are executed concurrently via an asynchronous MLLM serving backend, while each trajectory proceeds synchronously in a turn-based manner, with the agent awaiting tool response at every step.
Each trajectory consists of a sequence of reasoning and planning steps, tool invocations, tool responses, and a final answer.

\textbf{Reward Computation.} We then compute a reward $R_i$ for each trajectory $\tau_g$ that using a search-tool format reward $R_i^{format}$ and an outcome-level accuracy reward $R_i^{acc}$. (1) \textit{Format reward.}
The format reward encourages structured multi-turn reasoning and valid tool invocation by enforcing conformity to the expected interaction format.
Trajectories that adhere to the reasoning format $\tau$ defined in~\cref{Sec 3.2: DR-TTS} receive a reward of 1, whereas those that violate the format or produce unparsable tool-call queries receive a reward of 0.
(2) \textit{Accuracy reward.}
The accuracy reward evaluates whether the final answer produced by the model is correct.
Unlike previous methods that use rule-based evaluation, we extract the model's final answer
% enclosed within the $\textless$answer$\textgreater$ and $\textless$/answer$\textgreater$ tags
and assess its correctness using a powerful LLM by comparing it against the golden answer.
The final reward is defined as $R = \alpha R_{acc} + (1-\alpha) R_{format}$, where $\alpha$ balances the accuracy and format rewards. Subsequently, the group-relative advantage $A_g$ is computed according to the reward:
\begin{align}
    A_g = R_g - \frac{1}{G} \sum_{k=1}^{G} R_k .
\end{align}
\textbf{Optimization.} Finally, the GRPO employs a clipped objective with a KL penalty term to optimize the models:
% GRPO
\begin{align}
 &\mathcal{J}_{\mathrm{GRPO}}(\theta) = {} \mathbb{E}_{(I,T)\sim p_{\mathcal{D}},{\tau}\sim\pi_{\theta_\text{old}}(\cdot|I,T)}\nonumber\\
&\Biggl[ \frac{1}{G}\sum_{i=1}^{G} \min\ \!\Biggl(\frac{\pi_{\theta}({\tau}_i \mid I,T)}{\pi_{\theta_{\mathrm{old}}}({\tau}_i \mid I,T)}A_i, \mathrm{clip}\ \!\Bigl(\frac{\pi_{\theta}({\tau}_i \mid I,T)}{\pi_{\theta_{\mathrm{old}}}({\tau}_i \mid I,T)},\,\nonumber\\
&1-\epsilon,\,1+\epsilon\Bigr)A_i - \beta D_{\mathrm{KL}}\left(\pi_{\theta}| | \pi_{\mathrm{ref}}\right)\Biggr) \Biggr]\textrm{.}
\end{align}

\subsection{Evaluation Protocol}
At test time, we switch from the offline search tools to online search tools. Compared with offline search, online search often returns lengthy webpage content with substantial redundancy. Therefore, we employ an auxiliary LLM to verify and summarize tool responses before incorporating them into the reasoning process, preventing excessively long outputs from exceeding the model context limit and causing reasoning failures.

\begin{table*}[t]
  \centering
  % \vskip 0.1in
  % \vspace{-1em}
  \caption{\textbf{Main Results.} To examine the effectiveness of our method, we compare our MM-DeepResearch with other deep research agents, as well as powerful MLLMs under direct reasoning and RAG worlflow paradigm. $^{\dagger}$ denotes results evaluated by ourselves.}
  \scalebox{0.8}{
  \setlength{\tabcolsep}{5.5pt}
  \begin{tabular}{lccccccc}
    \toprule
    Method & SimpleVQA & MMSearch & LiveVQA & FVQA-test & InfoSeek & Browsecomp-VL & \makecell[c]{Average} \\
    \midrule
    \midrule
    % \textit{Direct Inference} & & & &  \\
    \multicolumn{8}{c}{\em Direct Reasoning}\\ \midrule
    GPT-4o~\cite{openai2024GPT-4o} & 51.7 & 18.7 & 28.1 & 48.0 & 52.9 & 5.5 & 34.1 \\
    GPT-5~\cite{GPT-5} & 54.1 & 35.1 & 44.4 & 54.4 & 61.7 & 48.6 & 49.7  \\
    % Gemini3 Pro & 75.3  \\
    Qwen3-VL-8B~\cite{Qwen3-VL} & 47.1 & 11.7 & 23.1 & 24.2  & 23.1 & 24.1 & 25.5 \\
    Qwen3-VL-32B~\cite{Qwen3-VL} & 58.0 & 19.8 & 45.5 & 34.1 & 28.0 & 30.8 & 36.0 \\
    % \textit{RAG Workflow} & & & &\\
    \midrule
    \midrule
    \multicolumn{8}{c}{\em RAG Workflow }\\ \midrule
    GPT-4o~\cite{openai2024GPT-4o} & 63.6 & 49.1 & 40.1 & 66.3 & 59.5 & 13.4 & 48.7 \\
    GPT-5~\cite{GPT-5} & 55.9 & 52.6 & 56.0 & 62.6 & 70.6 & 54.9 & 58.8 \\
    Qwen3-VL-8B~\cite{Qwen3-VL} & 62.3 & 47.3 & 39.3 & 53.6 & 46.1 & 29.3 & 46.3  \\
    % Qwen3-VL 32B~\cite{Qwen3-VL} &  &  &  \\
    \midrule
    \midrule
    \multicolumn{8}{c}{\em Agentic Search }\\ \midrule
    % \textit{Agentic MLLMs} & & & & \\
    % Qwen2.5-VL 7B~\cite{Qwen3-VL} & \\
    DeepMMSearch-R1-7B~\cite{narayan2025deepmmsearch} & 55.8 & -- & -- & -- & 47.5 & -- & --  \\
    Visual-ARFT-7B~\cite{liu2025visualarft} & 42.4 & 34.5 & 25.4 & 41.7 & 37.9 & 16.5$^{\dagger}$ & 33.1 \\
    MMSearch-R1-7B~\cite{wu2025mmsearchr1} & 57.4 & 53.8 & 48.4 & 58.4 & 55.1 & 20.9$^{\dagger}$ & 49.0 \\
    DeepEyes-v2-7B~\cite{hong2025deepeyesv2} & 59.4 & 63.7 & -- & 60.6 & 51.1 & -- & --  \\
    WebWatcher-7B~\cite{geng2025webwatcher} & 54.3 & 49.1 & 51.2 & -- & -- & 21.2 & --  \\
    \rowcolor{mygray} 
    MM-DeepResearch-7B & 62.0 & 61.4 & 60.0 & 61.9 & 58.7 & 32.8 & 56.1 \\
    \midrule
    Qwen3-VL-8B~\cite{Qwen3-VL} & 52.0 & 37.4 & 50.6 & 58.7 & 50.3 & 27.9 & 46.2 \\
    SenseNova-MARS-8B~\cite{chng2025sensenova} & 61.7 & 67.4 & 56.2 & 67.1 & 70.1 & 35.1$^{\dagger}$ & 59.6 \\
    \rowcolor{mygray} 
    MM-DeepResearch-8B & 65.9 & 67.8 & 65.0 & 69.2 & 72.3 & 37.9 & 63.0 \\
    \midrule
    % \multicolumn{9}{c}{\small \em Scaling to Larger Models }\\ \midrule
    Qwen3-VL-32B~\cite{Qwen3-VL} & 58.7 & 44.4 & 45.5 & 60.2 & 58.5 & 35.1 & 50.4 \\
    WebWatcher-32B~\cite{geng2025webwatcher} & 59.0 & 55.3 & 58.7 & -- & -- & 27.0 & --  \\
    \rowcolor{mygray} 
    MM-DeepResearch 32B & \textbf{67.6} & \textbf{69.0} & \textbf{68.0} & \textbf{70.1} & \textbf{73.9} & \textbf{43.0} & \textbf{65.3} \\
    \bottomrule
  \end{tabular}}
  \label{tab:Main Results}
  % \vskip -0.2in
  \vskip -0.1in
\end{table*}

\section{Experiment}

\subsection{Dataset.}
\textbf{The Source of Hyper-Search 3K.}
To construct a comprehensive QA dataset, we collect visual sources from 7 diverse categories (\ie, arts, sports, education, history, movies, places, and technology), as starting image nodes for Hyper-Search, ensuring broad domain coverage.

\textbf{Training Dataset.}
We prepare training data for SFT and RL stages as follows.
(1) For SFT, we select InfoSeek~\cite{infoseek} for generating search tool trajectories.
Before tree search, we select QA pairs that are challenging for Qwen3-VL-32B, and ultimately synthesize 10K valid search trajectories by DR-TTS for SFT.
(2) For RL, we use 3K samples from Hyper-Search and an additional 3K search-intensive samples selected from FVQA~\cite{wu2025mmsearchr1}.
% Additional details are provided in the Appendix.

\textbf{Evaluation Dataset}. We evaluate and compare MM-DeepResearch on six information-intensive benchmarks that require invoking search tools to solve. Detailed dataset descriptions are provided in the Appendix.

\subsection{Implementation Details}

\textbf{Search Engine.}
We detail the implementation of both online and offline search tools in this work.
(1) \textit{Online search.}
We employ SerpAPI
\footnote{SerpAPI: https://serpapi.com/}
as the online search engine for Hyper-Search and DR-TTS to retrieve both textual and visual information from the web, owing to its robust and versatile search functionality.
For image retrieval, we directly download the corresponding image files from the provided image URLs.
For textual retrieval, since SerpAPI returns a list of web URLs, we further use Jina Reader
\footnote{Jina Reader: https://jina.ai/reader/}
to fetch and parse the content of the associated web pages.
\textit{(2) Offline Search.}
We pre-collect multimodal web information and build an offline search engine that supports both text-based and multimodal retrieval.
Text-based retrieval is performed using a text-only E5 embedding model~\citep{wang2022e5_base_V2}, while multimodal retrieval is enabled by Jina-CLIP embeddings~\citep{gunther2025jina_embeddings_v4}.
The retrieval index is constructed using FlashRAG~\citep{jin2025flashrag}.

\textbf{Base Model.}
To examine the effectiveness of our search-intensive QA data, search trajectories, and offline search engine–based training, we select two kinds of MLLMs:
(1) MLLMs without native agentic tool-use capabilities (\ie, Qwen2.5-VL-7B-Instruct), which are used to assess whether our method can equip models with agentic search capabilities from scratch;
(2) MLLMs with native agentic tool-use capabilities (\ie, Qwen3-VL-8B and 32B-Instruct), which are used to examine whether our approach yields further performance gains beyond existing baselines.

\textbf{Training setup.}
For SFT, we use LLaMA-Factory~\cite{zheng2024llamafactory} to fully fine-tune the model using a batch size of 128, a learning rate of 5e-6, and training over 3 epochs. 
For RL, we use VeRL~\cite{sheng2024verl} with a global batch size of 128, 5 rollouts per prompt, and a learning rate of 1e-6.
The maximum number of tool invocations is capped at 5 per trajectory. To handle long multimodal tool responses and enhance long-context information integration capabilities, we scale the maximum training context length to 70,000 tokens.
For smaller MLLMs (\eg, 7B and 8B), both SFT and RL are conducted on 8 NVIDIA H100 GPUs, while larger models (\eg, 32B) are trained on 32 NVIDIA H100 GPUs.
In addition, we allocate 8 NVIDIA H100 GPUs to deploy the offline search engine and judge model (\ie, Qwen3-Next-80B-A3B-Instruct~\cite{qwen3technicalreport}).

\subsection{Main Results}
To evaluate our method, we conduct experiments on models without and with native agentic tool-use capabilities, and compare our MM-DeepResearch against various SOTAs, including non-agentic and agentic search MLLMs in~\cref{tab:Main Results}.

\textbf{Compared with non-agentic models.}
We first conduct experiments on Qwen2.5-VL-7B, which lacks native tool-calling capabilities.
Trained on our synthesized search trajectories and generated QA data, MM-DeepResearch-7B learns agentic search capabilities, enabling multi-step reasoning and the coordinated use of diverse tools to retrieve the required information and synthesize final answers, which significantly improves its performance on deep research tasks.
Under the same base model, MM-DeepResearch-7B outperforms prior agentic search MLLMs, i.e., Visual-ARFT and MMSearch-R1-7B, with average improvements of 23\% and 7.1\% across six benchmarks. Moreover, it surpasses the previous 7B state-of-the-art WebWatcher by 7.7\% and 12.3\% on SimpleVQA and MM-Search, respectively.
These results demonstrate the effectiveness of our approach in incentivizing agentic search capabilities from scratch.

\textbf{Compared with agentic models.}
We further evaluate our approach on agentic foundational MLLMs (i.e., Qwen3-VL-8B and Qwen3-VL-32B), which possess basic tool-use capabilities, to assess whether our approach can further enhance their performance.
As shown in~\cref{tab:Main Results}, MM-DeepResearch-8B averagely improves upon baseline Qwen3-VL-8B by 17\%.
Compared with deep research agent SenseNova-MARS-8B, MM-DeepResearch-8B achieves an average gain of 3.4 points, with improvements of 4.2\% on SimpleVQA.
In addition, when scaling the model to 32B, MM-DeepResearch-32B exhibits consistent performance gains of 14.9\% over baseline Qwen3-VL-32B, further validating its effectiveness in enhancing agentic foundational MLLMs.

\subsection{Ablation Study}
\textbf{Ablation study on Hyper-Search.}
We analyze how different data generation methods affect search depth (\ie, average tool calls) and performance in~\cref{tab: Ablation study on Hyper-Search}.
% 我们通过比较用不同数据训练后的结果进行RL训练by InfoSeek, graph-based, and Hyper-Search data, each containing 3K samples and initialized from our SFT model.
We compare RL training results using different datasets, \ie, InfoSeek, graph-based, and Hyper-Search, each containing 3K samples and initialized from our SFT model.
% RL training is performed on InfoSeek, graph-based, and Hyper-Search data, each containing 3K samples and initialized from our SFT model.
The results show that RL training on data constructed via Hyper-Search enables deeper exploratory behavior, encouraging more search turns and leading to more accurate answers grounded in more comprehensive information.

\begin{table}[t]
  \centering
    \caption{\textbf{Ablation Study on Hyper-Search. }
  }
  % \belowrulesep=0pt
% \aboverulesep=0pt
  % \vskip 0.1in
  % \vspace{-1em}
  \scalebox{0.8}{
  \setlength{\tabcolsep}{7pt}
  \begin{tabular}{l|cc}
 \toprule
 \multicolumn{1}{l|}{Method} & Avg. Tool Calls & MMSearch \\
% \hline
\midrule
    InfoSeek & 1.6 & 62.1 \\
    Graph-based & 1.7 & 64.8 \\
    Hyper-Search & 2.3 & 67.8 \\
    \bottomrule
  \end{tabular}}
  \label{tab: Ablation study on Hyper-Search}
  % \vskip -0.2in
  \vskip -0.05in
\end{table}

\begin{table}[t]
  \centering
    \caption{\textbf{Ablation study on MM-DeepResearch.}
  }
  % \vskip 0.1in
  % \vspace{-1em}
  \scalebox{0.76}{
  \setlength{\tabcolsep}{3.pt}
  \begin{tabular}{@{}l|cc|ccccccl@{}}
    \toprule
    Method & DR-TTS (SFT) & Hyper-Search (RL) & MMSearch \\
    \midrule
    Qwen3-VL (Baseline) & -- & --  &  37.4  \\
    MM-DeepResearch-SFT  & \CheckmarkBold & -- &  52.3  \\
    MM-DeepResearch-RL  & -- & \CheckmarkBold & 65.2 \\
    MM-DeepResearch & \CheckmarkBold & \CheckmarkBold & 67.8 \\
    \bottomrule
  \end{tabular}}
  \label{tab: ablation study on MM-DeepResearch}
  \vskip -0.1in
\end{table}

\textbf{Ablation study of MM-DeepResearch.}
We study the contributions of different components in MM-DeepResearch in~\cref{tab: ablation study on MM-DeepResearch}.
Starting from the Qwen3-VL baseline (37.4 on MMSearch), training with DR-TTS-explored search trajectories for SFT substantially improves performance to 52.3, indicating that our search trajectory data effectively enhances search behavior and information integration.
When trained only with Hyper-Search data using RL, MM-DeepResearch-RL further boosts performance to 65.2, demonstrating that the dataset incentivize multi-turn agentic search.
By jointly leveraging DR-TTS trajectory data for SFT and Hyper-Search QA data for RL, MM-DeepResearch achieves the best performance, reaching 67.8 on MMSearch, validating the effectiveness of our proposed methods.

\textbf{Ablation study on Search Tools.}
\cref{tab: ablation study on different search tools} presents an ablation study of offline search tools. Enabling information-based search progressively improves MMSearch performance, with T2T, T2I, and I2T increasing the score from 11.7 to 66.9.
Adding knowledge-based T2T search further boosts performance to 67.8, demonstrating the complementary effects of information-based and knowledge-based search in multimodal reasoning.

\begin{table}[t]
  \centering
    \caption{\textbf{Ablation Study of offline search tools.}
  }
  % \vskip 0.1in
  % \vspace{-1em}
  \scalebox{0.78}{
  \setlength{\tabcolsep}{10pt}
  \begin{tabular}{cccc|cccccl}
    \toprule
    \multicolumn{3}{c}{Information-based} & \multicolumn{1}{c}{Knowledge-based} & \multirow{2}{*}{MMSearch} \\
    \cmidrule(lr){1-3} \cmidrule(lr){4-4}
    % \cmidrule(lr){2-3} \cmidrule(lr){4-5}
    \multirow{1}{*}{T2T} & \multirow{1}{*}{T2I} & \multirow{1}{*}{I2T} & \multirow{1}{*}{T2T} &  \\
    \midrule
     -- & -- & -- & -- & 11.7  \\
    \CheckmarkBold & -- & --  & -- & 55.9 \\ 
    \CheckmarkBold & \CheckmarkBold & -- & -- & 64.7 \\
    \CheckmarkBold & \CheckmarkBold & \CheckmarkBold & -- & 66.9 \\
    \CheckmarkBold & \CheckmarkBold & \CheckmarkBold & \CheckmarkBold & 67.8  \\
    \bottomrule
  \end{tabular}}
  \label{tab: ablation study on different search tools}
  % \vskip -0.05in
\end{table}

\section{Conclusion}
This paper presents MM-DeepResearch, an agent designed for deep research tasks that integrates reasoning and planning, multi-turn search tool invocation, and long-context cross-modal information integration.
To develop MM-DeepResearch, we introduce three core components: Hyper-Search for search-intensive QA generation, DR-TTS for synthesizing effective search trajectories, and an offline search engine for scalable RL training.
Together, these designs enable training a search agent from scratch and substantially enhance its search and reasoning capabilities.
Extensive experiments demonstrate the effectiveness of our approach across models with and without native agentic capabilities.
We hope this work provides a simple yet effective baseline for future research on multimodal deep research agents.

% references without citing it in the main text, use \
\nocite{xiao2026mimov2-flash,team2025kimik2,wang2025step3,guo2025seed1_5vl,huang2026step3vl,intellect3,zhang2025web_search_agent,team2026longcat_flash_thinking}
\nocite{liang2025reasoning_rag,xia2025improving_rag,wei2025alignrag,li2025searcho1,jin2025searchr1}
\nocite{wang2026deepresearcheval,yao2024denseconnector,yao2024mulberry,yao2025r1-sharevl,yao2025mmreason,li2025mm-BrowseComp,compagnoni2025reag}

% \newpage
% \clearpage

% \section*{Impact Statement}
% This paper presents work whose goal is to advance the field of Multimodal Large Language Models and Agentic AI. There are many potential societal consequences of our work, none
% which we feel must be specifically highlighted here.

\bibliography{example_paper}
\bibliographystyle{icml2026}

%%%%%%%%%%%%%%%%%%%%%%%%%%%%%%%%%%%%%%%%%%%%%%%%%%%%%%%%%%%%%%%%%%%%%%%%%%%%%%%
%%%%%%%%%%%%%%%%%%%%%%%%%%%%%%%%%%%%%%%%%%%%%%%%%%%%%%%%%%%%%%%%%%%%%%%%%%%%%%%
% APPENDIX
%%%%%%%%%%%%%%%%%%%%%%%%%%%%%%%%%%%%%%%%%%%%%%%%%%%%%%%%%%%%%%%%%%%%%%%%%%%%%%%
%%%%%%%%%%%%%%%%%%%%%%%%%%%%%%%%%%%%%%%%%%%%%%%%%%%%%%%%%%%%%%%%%%%%%%%%%%%%%%%
\newpage
\appendix
\onecolumn
\section*{Appendix}

\section{Multimodal QA Dataset}
\begin{figure}[htbp]
\centering
\includegraphics[width=0.75\linewidth]{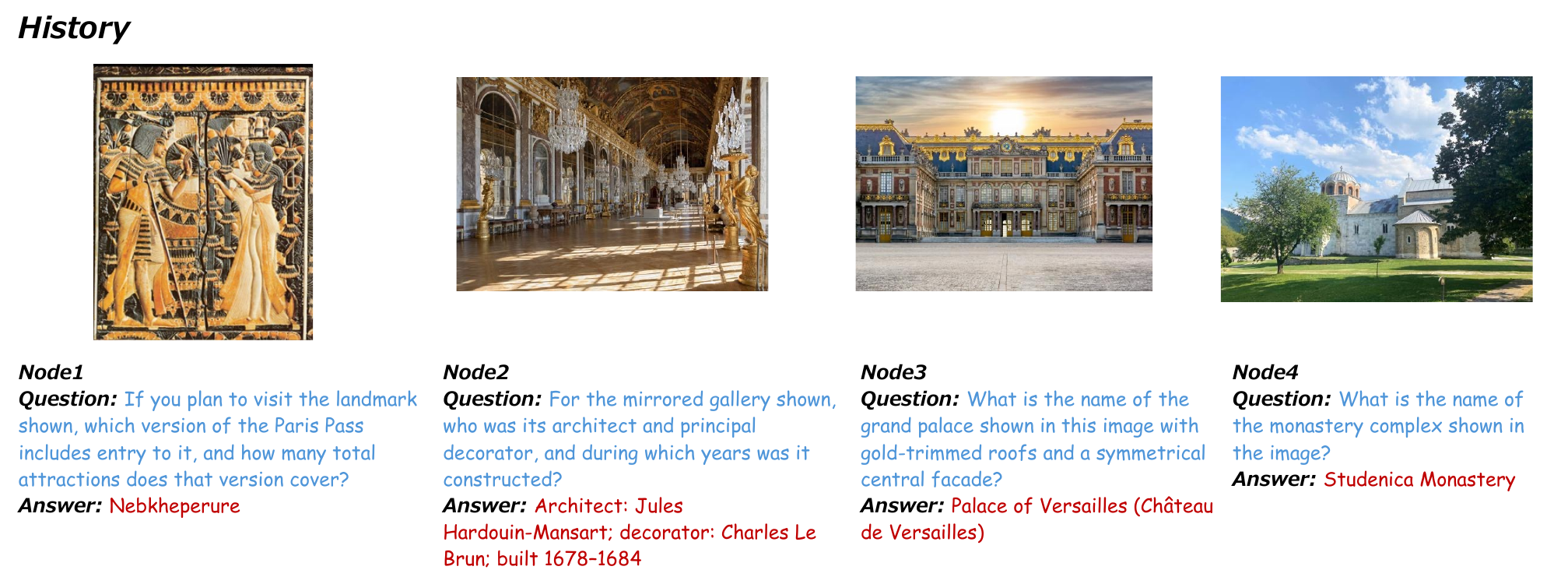}
\caption{
Data examples from the history category.}
\label{fig:history}
\end{figure}

\begin{figure}[t]
\centering
\includegraphics[width=0.75\linewidth]{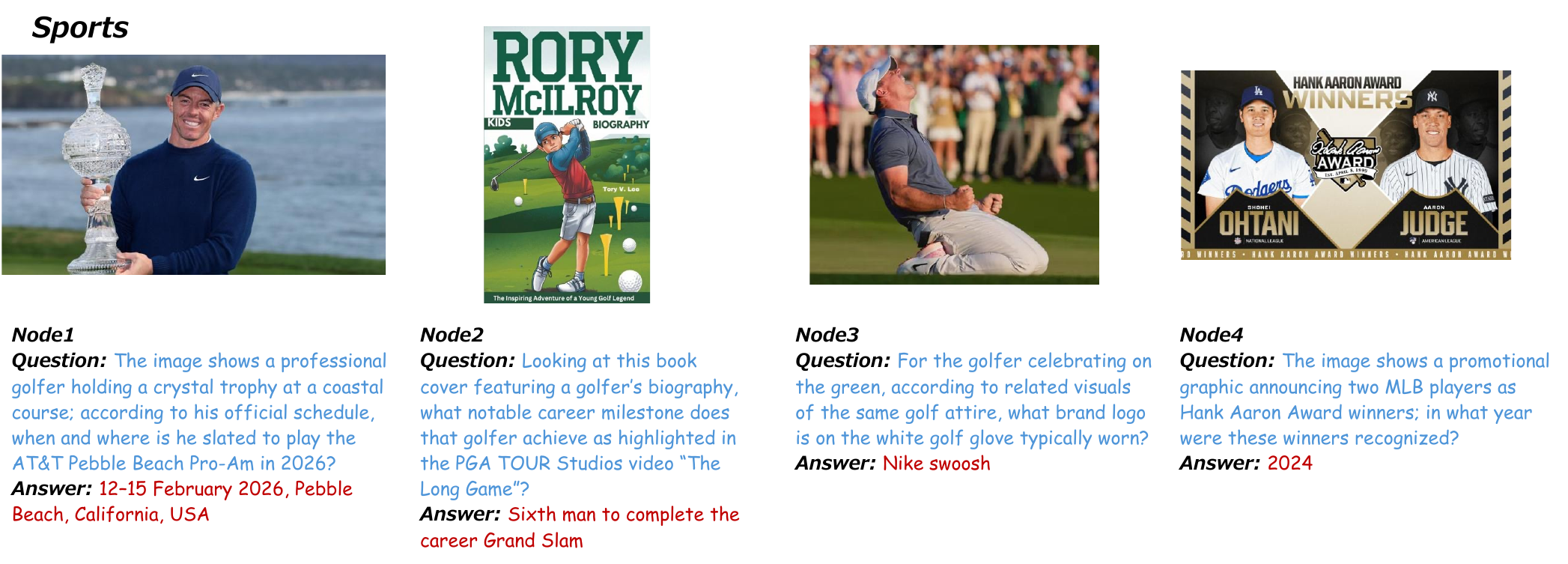}
\caption{
Data examples from the sports category.
}
\label{fig:sports}
\end{figure}

To construct a comprehensive QA dataset, we collect visual sources from seven diverse categories (\ie, arts, sports, education, history, movies, places, and technology), as starting image nodes for Hyper-Search, ensuring broad domain coverage. We provide some examples of our Multimodal QA Dataset as shown in Fig.~\ref{fig:history} $\sim$ Fig.~\ref{fig:sports},
% we try to cover different QA categories in every task to offer a holistic overview of multimodal QA Dataset.

\begin{figure}[htbp]
\centering
\includegraphics[width=0.75\linewidth]{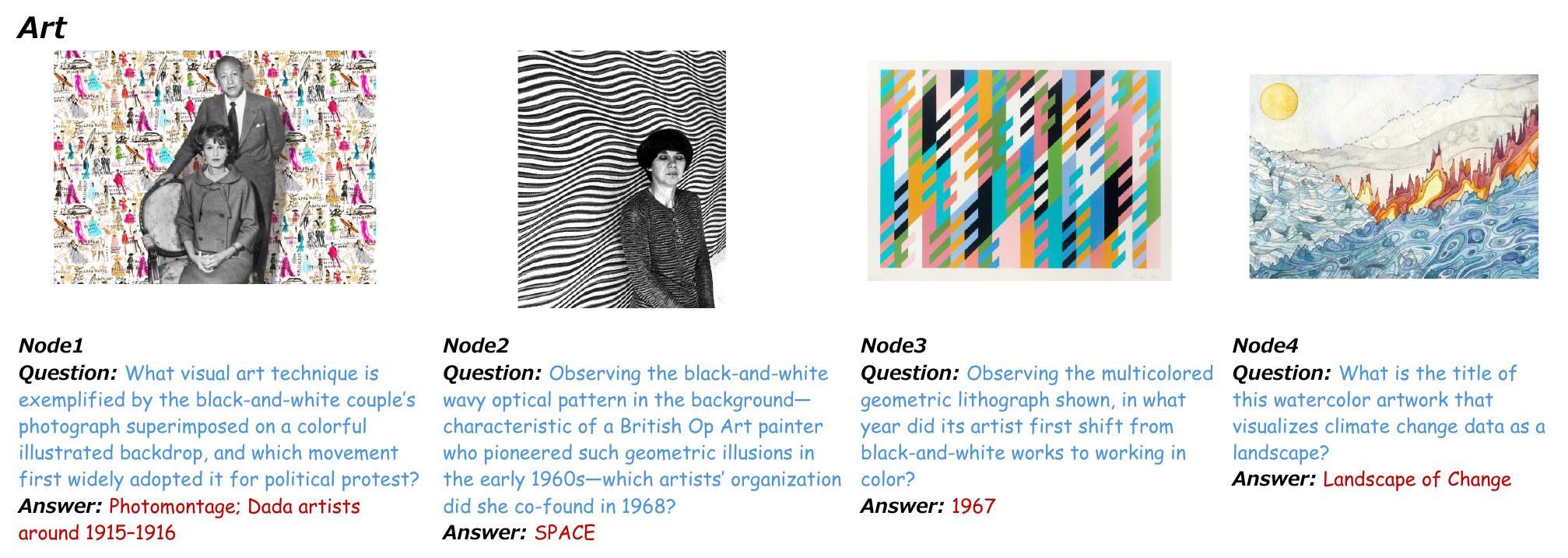}
\caption{
Data examples from the art category.
}
\label{fig:art}
\end{figure}

\begin{figure}[ht]
\centering
\includegraphics[width=0.75\linewidth]{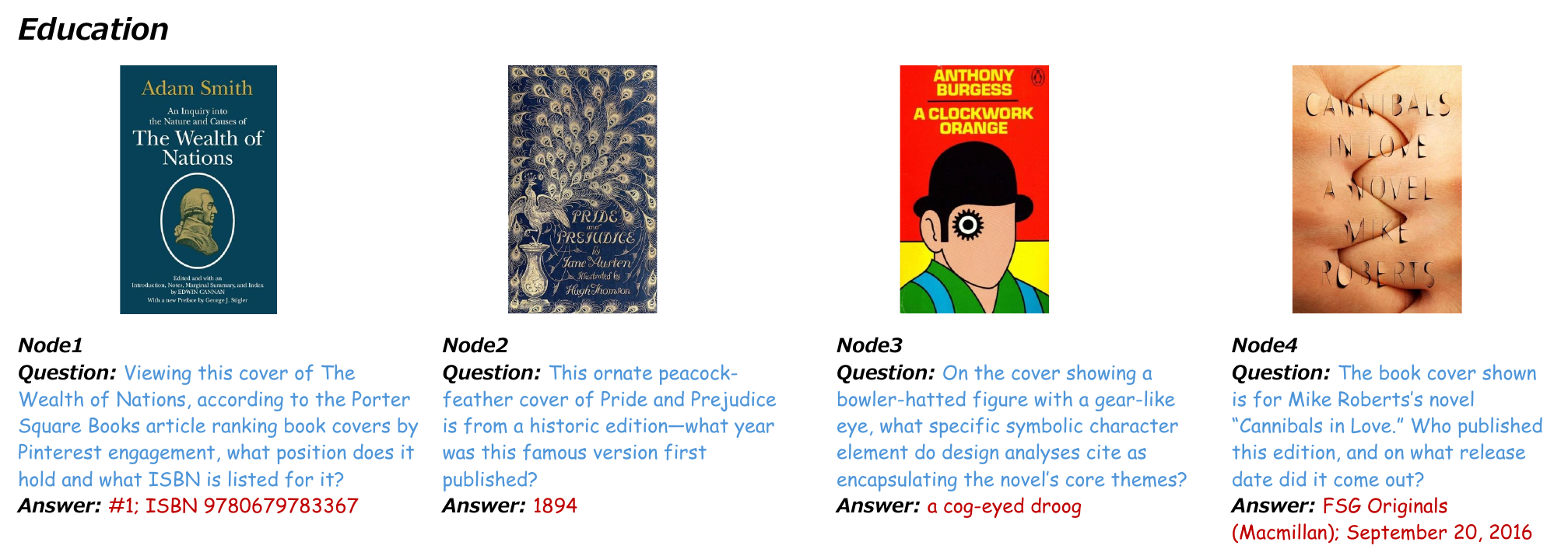}
\caption{
Data examples from the education category.
}
\label{fig:education}
\end{figure}

\begin{figure}[t]
\centering
\includegraphics[width=0.75\linewidth]{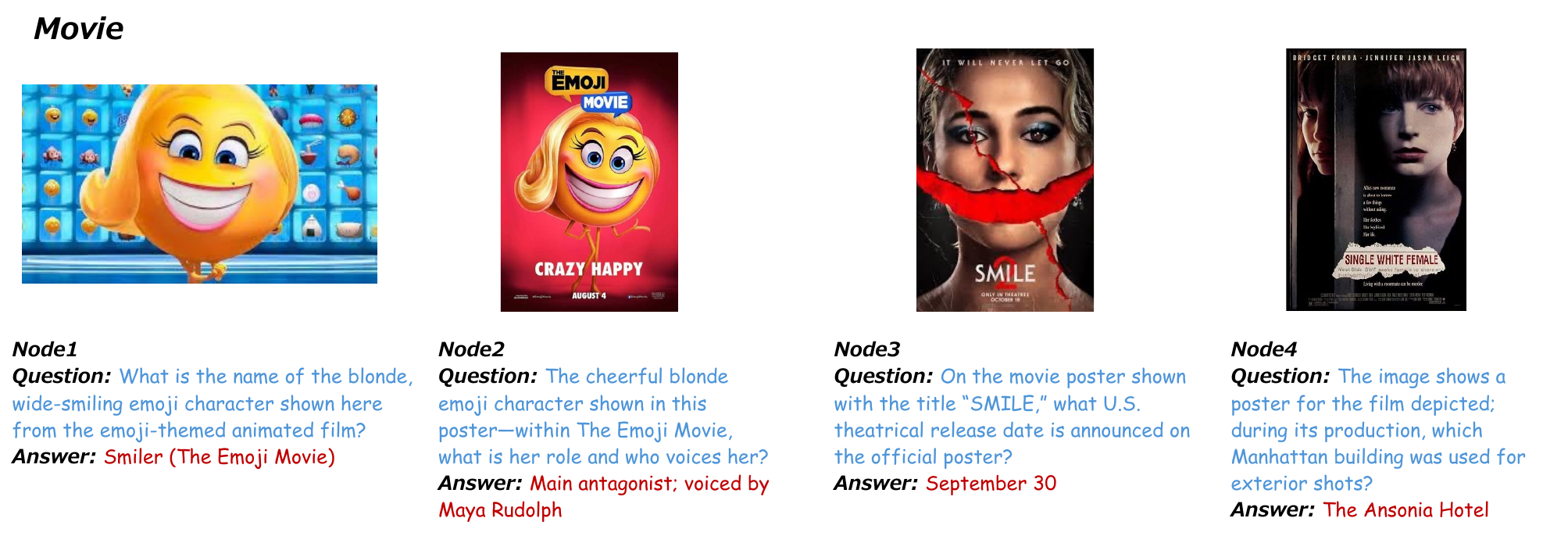}
\caption{
Data examples from the movie category.
}
\label{fig:movie}
\end{figure}

\begin{figure}[t]
\centering
\includegraphics[width=0.75\linewidth]{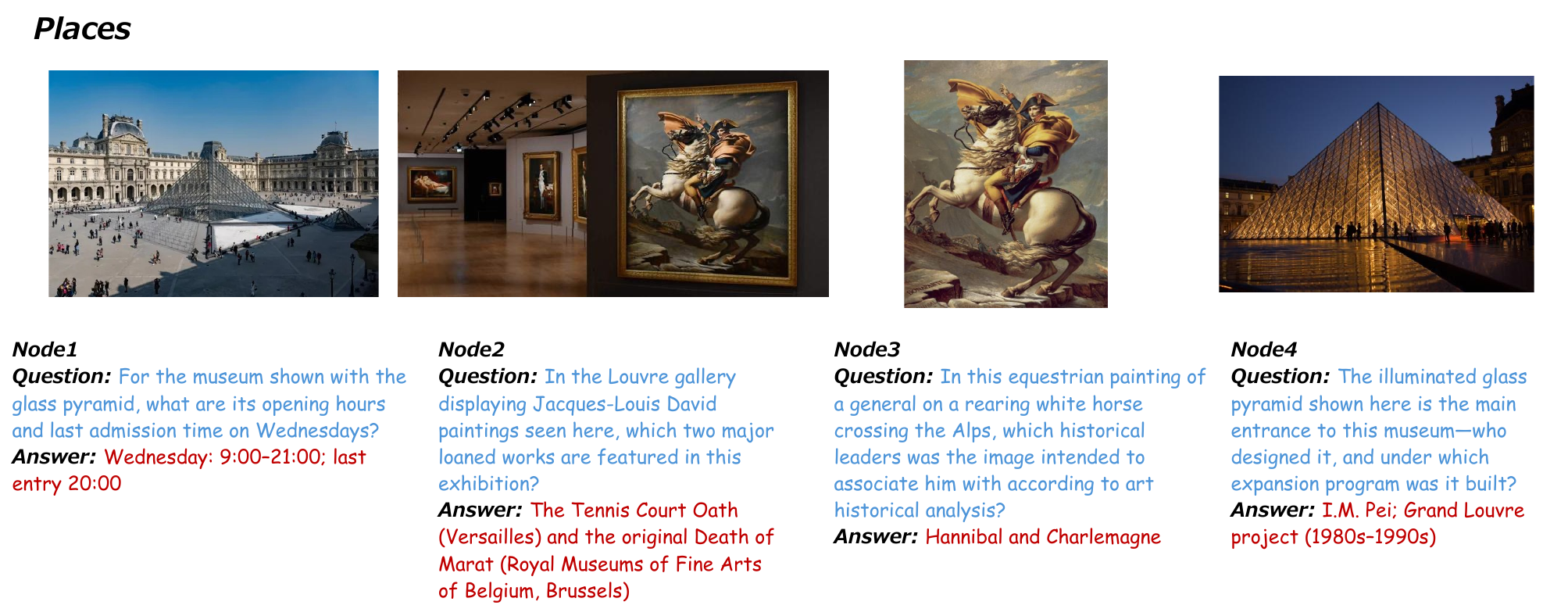}
\caption{
Data examples from the places category.
}
\label{fig:places}
\end{figure}

\begin{figure}[t]
\centering
\includegraphics[width=0.75\linewidth]{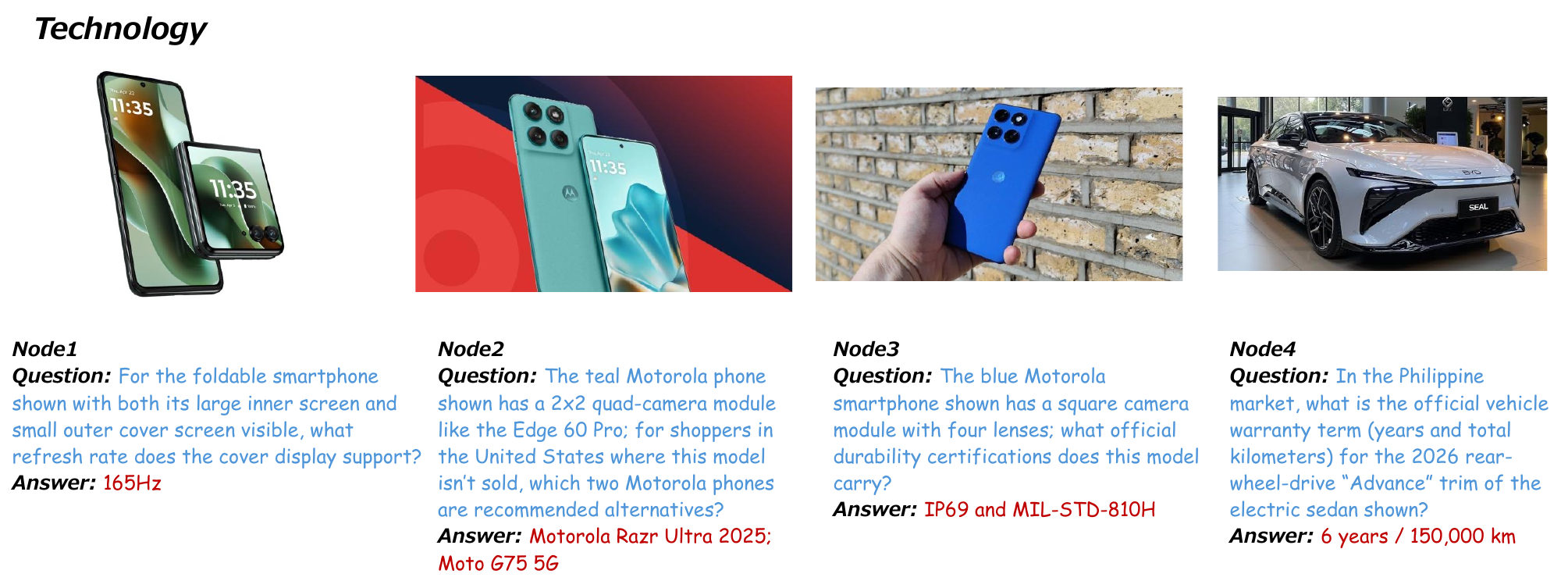}
\caption{
Data examples from the technology category.
}
\label{fig:technology}
\end{figure}

\newpage
\clearpage

\section{More Setup}

\textbf{Offline Search Engine Implementation.}
To enable offline corpus search, we pre-collect a multimodal corpus and adopt a retrieval framework based on FlashRAG~\citep{jin2025flashrag}.
For the textual corpus, we prompt GPT to generate multiple candidate search queries for each question and use SerpAPI\footnote{SerpAPI: https://serpapi.com/} and Jina\footnote{Jina Reader: https://jina.ai/reader/} to retrieve the corresponding webpage content.
In addition, we incorporate large-scale English Wikipedia (2018) to improve corpus coverage and diversity, simulating the large-scale information available on the web.
These web contents are segmented into 512-word passages with titles appended and employ e5-base-v2~\citep{wang2022e5_base_V2} as the text retriever.
For the image corpus, we additionally prompt GPT to generate multiple candidate image search queries and use SerpAPI to retrieve relevant image URLs. We employ jina-embeddings-v4~\citep{gunther2025jina_embeddings_v4} as the multimodal retriever to enable cross-modal retrieval.

\textbf{Hyperparameters of Search Engine.}
During training, we retrieve the top-3 most similar textual passages for each query. For image retrieval, only images with an embedding similarity greater than 0.7 are retained, and for each tool call query we keep only the single most similar image to avoid excessively long reasoning contexts and excessive memory consumption.
During evaluation, we adopt online retrieval, expanding textual search to the top-5 results and retrieving up to three images per query.

% \textbf{Model Implementation.} In Hyper-Search, we use Qwen3-VL-235B-A22B-Instruct 

\section{Evaluation Benchmark}
We evaluate MM-DeepResearch across six benchmarks, which are described as follows.
\begin{itemize}
    \item \textbf{SimpleVQA.} SimpleVQA~\citep{cheng2025simplevqa} is a newly proposed multimodal benchmark designed to evaluate the factuality of MLLMs when answering short natural-language questions grounded in images. It encompasses diverse tasks and scenarios, with high quality and challenging queries aligned across multiple topics. The benchmark ensures static, verifiable reference answers and employs rigorous quality controls, making it straightforward to evaluate model outputs with minimal variance. Using a combination of manual verification and an LLM as a judge scoring system, SimpleVQA facilitates comprehensive empirical comparisons across leading MLLMs and text only LLMs, providing insight into image comprehension, generative accuracy, and common failure modes.
    Following WebWatcher~\cite{geng2025webwatcher}, we evaluate on the same 300 examples sampled from the 1,013 English QA pairs.
    \item \textbf{MMSearch.} MMSearch~\citep{jiang2024mmsearch} represents a multimodal browsing benchmark crafted to place rigorous demands on models that must integrate visual and textual signals with external retrieval to answer complex queries. Unlike prior multimodal tasks where salient objects may be easily recognizable, this benchmark requires exhaustive reasoning over localized visual cues, disciplined cross modal evidence gathering, provenance verification under noisy search results, and long, tool augmented reasoning chains.
    Among them, 171 examples include images, and we evaluate on these 171 image-based examples to align with previous methods.
    \item \textbf{LiveVQA.} LiveVQA~\citep{fu2025livevqa} is a large scale visual question answering dataset that focuses on up to date visual knowledge. It contains over 107,000 samples across multiple real world categories sourced from recent news, video, and academic platforms, making it a benchmark tailored for evaluating how MLLMs handle real time visual information beyond their training cutoffs. The dataset supports studies on both performance gaps in current models and parameter efficient fine tuning strategies that balance new visual knowledge integration with core perception and reasoning capabilities.
    We also evaluate on a 300-example subset sampled from WebWatcher.
    \item \textbf{FVQA.} FVQA~\citep{wang2017fvqa} is a foundational visual question answering dataset that extends traditional VQA tasks by requiring external commonsense knowledge to answer visual questions correctly. In FVQA, each image–question–answer triplet is accompanied by a supporting fact represented as a structured knowledge triplet, drawn from common sense knowledge bases. This design forces models to reason beyond visual perception and to retrieve and integrate factual knowledge, thus serving as a core benchmark for knowledge based VQA research involving explicit external knowledge grounding.
    We test on the full set of 1,800 high-quality examples here.
    \item \textbf{InfoSeek.} InfoSeek~\cite{infoseek} is a visual question answering dataset that concentrates on information seeking questions – queries that cannot be resolved using only common sense or visual content presented in the image. The dataset blends human annotated and large scale automatically generated question–answer pairs, challenging models to leverage fine grained knowledge beyond typical VQA tasks. Evaluation using InfoSeek highlights the limitations of current pre trained vision language models in answering knowledge intensive questions, and underscores the performance gains enabled by fine tuning with explicit information seeking objectives.
    Here, we use 2,000 instances for evaluation, sampled by MMSearch-R1~\cite{wu2025mmsearchr1}. 
    \item \textbf{Browsecomp-VL.} BrowseComp-VL~\citep{geng2025webwatcher} is a recently introduced benchmark that extends BrowseComp style tasks into the multimodal domain, requiring strategic planning and cross modal retrieval across text, images, and web content. Designed to evaluate sophisticated multimodal agents, BrowseComp-VL demands deep search, sequential evidence synthesis, and multimodal reasoning. By embedding both visual and textual clues into long horizon information seeking problems, it assesses agents’ ability to coordinate perception, retrieval, and planning in complex real world scenarios, and serves as an evaluation frontier for integrated vision language deep research.
    We evaluate on the full BrowseComp-VL dataset.
\end{itemize}

\section{More Discussion}

\subsection{Discussion on Different Embedding Models for the Offline Search Engine}
We investigate the impact of different embedding models for offline text retrieval, with results reported in Table~\ref{tab:Discussion on Different Embedding Models for the Offline Search Engine.}.
Overall, the performance differences between the two embedding models are marginal, indicating that offline text retrieval is relatively robust to the choice of embedding model.
Although jina-embeddings-v4~\cite{gunther2025jina_embeddings_v4} achieves slightly better performance, it incurs higher cost due to its high-dimensional embeddings and slower inference speed.
Considering efficiency and scalability, we therefore adopt e5-base-v2~\cite{wang2022e5_base_V2} for text embedding, which provides comparable retrieval quality while being significantly more efficient for large-scale offline indexing and retrieval.

\begin{table}[h]
  \centering
    \caption{\textbf{Discussion on Different Embedding Models for the Offline Search Engine.}
  }
  % \vskip 0.1in
  % \vspace{-1em}
  \scalebox{0.86}{
  \setlength{\tabcolsep}{3.5pt}
  \begin{tabular}{@{}lccccccccl@{}}
    \toprule
    Embedding Model & MMSearch  \\
    \midrule
    e5-base-v2 & 55.9 \\ 
    jina-embeddings-v4 & 56.1 \\ 
    \bottomrule
  \end{tabular}}
  \label{tab:Discussion on Different Embedding Models for the Offline Search Engine.}
  % \vskip -0.2in
\end{table}

\subsection{Comparing Online and Offline Search Performance at Evaluation}
We compare the performance of online and offline search engines during evaluation, as shown in Table~\ref{tab: discussion of Offline Text Search Engine}.
Overall, online search consistently achieves better performance, benefiting from access to up-to-date and large-scale web information.
Nevertheless, the offline search engine attains competitive results on both SimpleVQA and MMSearch, demonstrating its ability to retrieve informative and relevant evidence.
These results indicate that although offline search does not fully match online search performance, it is sufficiently effective to provide meaningful retrieval signals and can adequately stimulate the model to learn search behaviors.
This makes offline search a practical and cost-efficient alternative for training.

\begin{table}[h]
  \centering
    \caption{\textbf{Discussion of Offline Text Search Engine.}
  }
  % \vskip 0.1in
  % \vspace{-1em}
  \scalebox{0.86}{
  \setlength{\tabcolsep}{3.5pt}
  \begin{tabular}{@{}lccccccccl@{}}
    \toprule
    Search Engine & SimpleVQA & MMSearch \\
    \midrule
    Offline & 63.4 & 62.7  \\
    Online & 65.9 & 67.8  \\

    \bottomrule
  \end{tabular}}
  \label{tab: discussion of Offline Text Search Engine}
  % \vskip -0.2in
\end{table}

\subsection{Discussion on the Impact of Top-$k$ Text Retrieval during Training}
We analyze the effect of varying the number of retrieved textual passages ($k$) during training, with results shown in Table~\ref{tab:Discussion of Offline Text Search Engine}.
Overall, moderate values of $k$ lead to better performance, while increasing $k$ beyond a certain point yields diminishing returns.
Specifically, retrieving five passages achieves the best performance on MMSearch, outperforming both smaller ($k=3$) and larger ($k=10$) settings.
When $k$ becomes too large, the inclusion of excessive and potentially noisy information may dilute relevant evidence and increase reasoning difficulty, leading to performance degradation.

\begin{table}[h]
  \centering
    \caption{\textbf{Discussion of Offline Text Search Engine.}
  }
  % \vskip 0.1in
  % \vspace{-1em}
  \scalebox{0.86}{
  \setlength{\tabcolsep}{3.5pt}
  \begin{tabular}{@{}lccccccccl@{}}
    \toprule
    Embedding Model & Top-$k$ & MMSearch \\
    \midrule
    e5 & 3 & 55.9 \\ 
    e5 & 5 & 56.3 \\ 
    e5 & 10 & 54.8\\ 
    \bottomrule
  \end{tabular}}
  \label{tab:Discussion of Offline Text Search Engine}
  % \vskip -0.2in
\end{table}

\subsection{Discussion of Online and Offline Search Costs.}
\cref{tab: Ablation Study of Online and Offline Search Costs} compares training costs for online and offline search during RL.
Online search incurs substantial costs (around 640 dollars per hundred steps) due to repeated external API calls in each rollout and also introduces high latency, with an average response time of about 60 seconds per sample.
In contrast, offline search operates with no additional search-related cost and significantly lower latency (about 1 second per sample).
These results show that offline search is a cost-efficient and scalable alternative, particularly for GRPO settings with multiple rollouts and large-scale training.

\begin{table}[h]
  \centering
    \caption{\textbf{Discussion training cost and efficiency for online and offline search.}
  }
  % \vskip 0.1in
  % \vspace{-1em}
  \scalebox{0.78}{
  \setlength{\tabcolsep}{8pt}
  \begin{tabular}{@{}l|ccccccccl@{}}
    \toprule
    Method & Training Cost ($\$$) & Training Time (s / sample) \\
    \midrule
    Online Search & 640 & 60 \\
    Offline Search & 0 & 1 \\
    \bottomrule
  \end{tabular}}
  \label{tab: Ablation Study of Online and Offline Search Costs}
  % \vskip -0.2in
  \vskip -0.1in
\end{table}

\section{Case Study}
\begin{figure*}[htbp]
\centering
\includegraphics[width=0.9\linewidth]{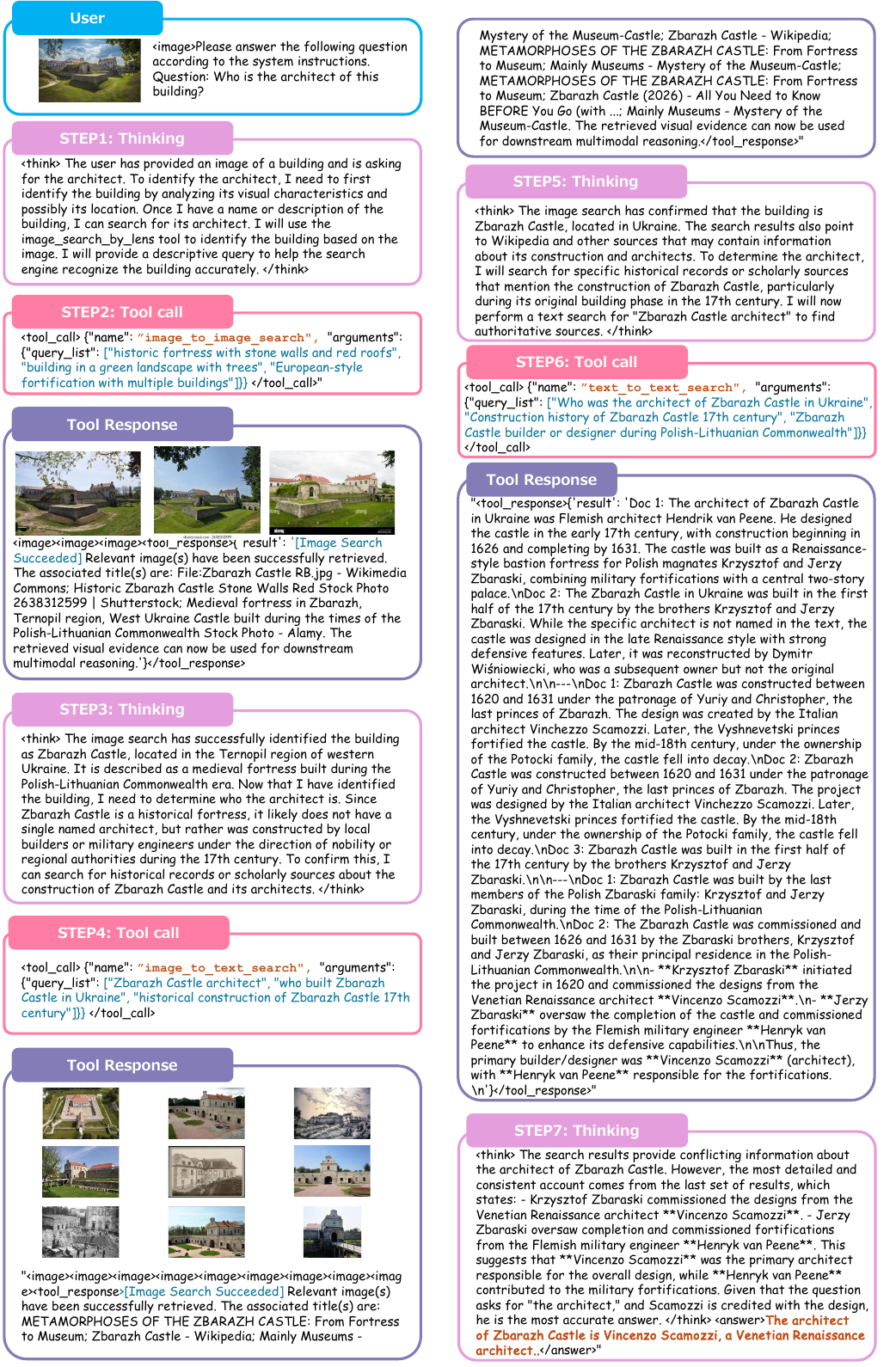}
\caption{
Case study of MM-DeepResearch.
}
\label{fig:case_study}
\end{figure*}

We present the inference cases of MM-DeepResearch in Fig.~\ref{fig:case_study}. In the reasoning trajectory, MM-DeepResearch alternates between thinking and tool calling. Regarding tool usage, the process begins with the image-to-image
% (\ie, \verb|image_search_by_lens|)
search tool to retrieve corresponding web addresses and associated metadata from the original images. It then uses the image-to-text 
% (\ie, \verb|image_search_by_text_query|)
search tool to gather supplementary visual data for enhanced reasoning. Finally, the text-to-text
% (\ie, \verb|text_search|)
search tool is invoked to retrieve relevant textual information. Ultimately, through thinking, all the gathered information is integrated to derive the final answer.

\newpage
\section{PROMPT}
\subsection{Deep Research Prompt}
Full prompt used during training and inference for the MM-DeepResearch model.

\begin{tcolorbox}[
    title=\textbf{System Message},
    breakable
]
\lstset{
    language=Python,
    basicstyle=\small\fontfamily{zi4}\selectfont,
    breaklines=true,
    breakindent=0pt,   % ← 关键：去掉自动换行缩进
    columns=fullflexible,
    commentstyle=\color{black},
    keywordstyle=\color{black},
    frame=none,
    extendedchars=true,
    showstringspaces=false,
    inputencoding=utf8
}
\begin{lstlisting}[numbers=none]
You are a helpful and harmless deep research assistant. Your task is to think carefully, seek external information when necessary, and provide accurate, well-supported answer to the user's question.

# Think guidelines
1. Reason step by step to solve the user's question. Decompose the original question into clear, manageable sub-questions.
2. After each reasoning cycle, summarize what has been established so far and decide whether additional sub-questions or external information are required.
3. Your thinking process MUST remain internal and structured within <think>...</think>.

# Tool usage guidelines
1. Use tools when external information is required to answer the question accurately.
2. Tool queries must be specific and concrete. Avoid ambiguous references or pronouns (e.g., ''it'', ''this'', ''he''), and use explicit entity names, dates, technical terms, or unique identifiers.
3. Effective tool usage depends on formulating high-quality queries and extracting useful information from tool responses.
4. Enclose all tool calls within <tool_call>...</tool_call>, and all tool outputs within <tool_response>...</tool_response>.

# Answer guidelines
1. If no external information or detailed explanation is required, always provide a concrete final answer enclosed within <answer>...</answer> (e.g., <answer>Beijing</answer>).

# Format guidelines
The assistant may follow a valid execution path as follows:
<think>reasoning</think>
(If tool usage is required)
<tool_call>tool invocation</tool_call>
<tool_response>tool output</tool_response>
(The above steps may be repeated if necessary)
<think>final reasoning</think>
<answer>final answer</answer>

# Tools
You may call one or more functions to assist with the user query.
You are provided with function signatures within <tools></tools> XML tags:
<tools>
{"type": "function", "function": {"name": "image_search_by_text_query", "description": "Searches images on the web based on the given query and returns relevant image results with their associated titles. This tool should only be used once.", "parameters": {"type": "object", "properties": {"query_list": {"type": "array", "description": "A list of fully-formed semantic queries for image search. The tool retrieves relevant images for this query."}}, "required": ["query_list"]}}}
{"type": "function", "function": {"name": "image_search_by_lens", "description": "Performs an image search using the image from the original question, refined with complementary text queries, and returns relevant images with their associated titles. This tool should only be used once.", "parameters": {"type": "object", "properties": {"query_list": {"type": "array", "description": "A list of text queries to accompany the image search. The tool retrieves relevant images for this image."}}, "required": ["query_list"]}}}
{"type": "function", "function": {"name": "text_search", "description": "Searches the web for relevant information based on the given query.", "parameters": {"type": "object", "properties": {"query_list": {"type": "array", "description": "A list of fully-formed semantic queries. The tool will return search results for each query."}}, "required": ["query_list"]}}}
{"type": "function", "function": {"name": "model_search", "description": "Queries an expert model to answer questions based on the given query.", "parameters": {"type": "object", "properties": {"query_list": {"type": "array", "description": "A list of fully-formed semantic queries. The tool will return the response for each query."}}, "required": ["query_list"]}}}
</tools>

For each function call, return a json object with function name and arguments within <tool_call></tool_call> XML tags:
<tool_call>
{"name": <function-name>, "arguments": <args-json-object>}
</tool_call>
\end{lstlisting}
\end{tcolorbox}
%%%%%%%%%%%%%%%%%%%%%%%%%%%%%%%%%%%%%%%%%%%%%%%%%%%%%%%%%%%%%%%%%%%%%%%%%%%%%%%
%%%%%%%%%%%%%%%%%%%%%%%%%%%%%%%%%%%%%%%%%%%%%%%%%%%%%%%%%%%%%%%%%%%%%%%%%%%%%%% 将这个呈现的prompt的字体和排版美化一下 现在好丑

\begin{tcolorbox}[
    title=\textbf{Prompt},
    breakable
]
\lstset{
    language=Python,
   basicstyle=\small\fontfamily{zi4}\selectfont,
    breaklines=true,
    columns=fullflexible,
    commentstyle=\color{black},
    keywordstyle=\color{black},
    frame=none,
    extendedchars=true,
    showstringspaces=false,
    inputencoding=utf8
}
\begin{lstlisting}[numbers=none]
Please answer the following question according to the system instructions.
Question: 

\end{lstlisting}
\end{tcolorbox}

\subsection{RL reward judgement Prompt}
The full prompt used during the RL training phase to evaluate answer correctness and compute the reward.

%% RL训练阶段用于判断答案正确性，计算reward的完整prompt

\begin{tcolorbox}[
    title=\textbf{System Message},
    breakable
]
\lstset{
    language=Python,
   basicstyle=\small\fontfamily{zi4}\selectfont,
    breaklines=true,
    breakindent=0pt,   % ← 关键：去掉自动换行缩进
    columns=fullflexible,
    commentstyle=\color{black},
    keywordstyle=\color{black},
    frame=none,
    extendedchars=true,
    showstringspaces=false,
    inputencoding=utf8
}

\begin{lstlisting}[numbers=none]
You are an AI assistant tasked with evaluating the correctness of model responses based on the question, and ground truth answer.
Your judgment should follow these principles:
1. Consider the question, and ground truth answer holistically before evaluating the model's response.
2. Your decision should be strictly Yes or No, based on whether the model's response is factually accurate and aligns with the ground truth answer.
3. If the model response is a more specific form of the ground truth answer, it is correct.
4. If the model response includes all key information but adds minor details, it is correct as long as the extra details are factually correct.
5. If the model response contradicts, modifies, or omits critical parts of the answer, it is incorrect.
6. For numerical values, ensure correctness even when presented in different units.
7. For names, check for first and last name correctness. If the middle name is extra but correct, consider it correct.
8. For yes/no questions, the response must exactly match "Yes" or "No" to be correct.
9. If the model response contains refusal statements, and does not directly answer the question, it must be judged incorrect.
10. If there are multiple candidate answers, you can also evaluate the model's response against all of them. If the response aligns with at least one candidate according to the rules above, it should be considered correct.
Your output must be in the following format: Yes or No

\end{lstlisting}
\end{tcolorbox}

\begin{tcolorbox}[
    title=\textbf{Prompt},
    breakable
]
\lstset{
    language=Python,
   basicstyle=\small\fontfamily{zi4}\selectfont,
    breaklines=true,
    breakindent=0pt,   % ← 关键：去掉自动换行缩进
    columns=fullflexible,
    commentstyle=\color{black},
    keywordstyle=\color{black},
    frame=none,
    extendedchars=true,
    showstringspaces=false,
    inputencoding=utf8
}

\begin{lstlisting}[numbers=none]
Question, and Model Response Evaluation
Question: {question}
Ground Truth Answer: {ground_truth_answer}
Candidate Answers: {candidate_answers}
Model Response: {model_response}
Evaluation Instructions
Evaluate whether the Model Response is correct based on the Question, Ground Truth Answer and Candidate Answers.
Follow the predefined judgment rules and provide a clear Yes/No answer without any illustrations.
Output Format Yes or No


\end{lstlisting}
\end{tcolorbox}

\subsection{QA generation Prompt}
The complete prompt used to generate QA pairs for training data.

\begin{tcolorbox}[
    title=\textbf{QA generation Prompt},
    breakable
]
\lstset{
    language=Python,
   basicstyle=\small\fontfamily{zi4}\selectfont,
    breaklines=true,
    breakindent=0pt,   % ← 关键：去掉自动换行缩进
    columns=fullflexible,
    commentstyle=\color{black},
    keywordstyle=\color{black},
    frame=none,
    extendedchars=true,
    showstringspaces=false,
    inputencoding=utf8
}

\begin{lstlisting}[numbers=none]
You are a question-answer generation agent for search-intensive multimodal reasoning.

Your task is to generate ONE high-quality QUESTION-ANSWER pair based on a query image and external evidence.

Important constraints:
- The QUESTION must be DIRECTLY ABOUT what is shown in the query image.
- The QUESTION must NOT be answerable using the image alone.
- The ANSWER must be derived ONLY from the external evidence provided below,
  NOT from the query image itself.
  
You are given the following information:

[Query Image]
- An image is provided as the query.
- You must infer what the image depicts using visual cues only.
- Do NOT assume or invent any textual description of the query image.

[External Textual Evidence]
- Multiple summaries of webpages retrieved via search:
{TEXT_SUMMARIES}

[External Visual Evidence]
- Captions of other visually related images retrieved from the web:
{OTHER_IMAGE_CAPTIONS}

[Instructions for the QUESTION]
1. The question MUST explicitly refer to the content of the query image
   (e.g., the object, structure, location, scene, or event visible in the image).
2. The question MUST require factual knowledge, identification, or context
   that CANNOT be obtained from the image alone.
3. The question SHOULD naturally arise from observing the image
   and then seeking more information.
4. Avoid yes/no questions and avoid vague or subjective wording.
5. Do NOT mention image captions, summaries, search results, or external evidence explicitly.

[Instructions for the ANSWER]
6. The answer MUST be factual, concise, and directly answer the question.
7. The answer MUST be supported by the provided external textual or visual evidence.
8. The answer MUST NOT rely on assumptions or information visible only in the query image.
9. Prefer answers that require synthesizing information from multiple pieces of evidence.
10. The answer MUST NOT be written as a complete grammatical sentence.

[Output Format]
Output exactly the following JSON object and nothing else:

{{
    "question": "<one clear, well-formed question>",
    "answer": "<a concise, factual answer>"
}}


\end{lstlisting}
\end{tcolorbox}

\end{document}